\documentclass[runningheads]{llncs}
\usepackage[T1]{fontenc}
\usepackage{graphicx}
\usepackage{amsmath,amssymb}
\usepackage{csquotes,booktabs}
\usepackage{multirow,array}
\usepackage{orcidlink}

\newcommand{\para}[1]{\vspace{0.5em} \noindent \textbf{#1} \hspace{0.5em}}

\begin{document}
\title{ID2image: Leakage of non-ID information \\ into face descriptors and inversion \\ from descriptors to images}
\titlerunning{ID2image}
%


\author{Mingrui Li\orcidlink{0000-0002-9504-7084} \and
William~A.~P.~Smith\orcidlink{0000-0002-6047-0413} \and
Patrik Huber\orcidlink{0000-0002-1474-1040}}

\authorrunning{M.~Li et al.}

\institute{University of York, York YO10 5DD, UK \\
\email{\{ml1652,william.smith,patrik.huber\}@york.ac.uk}
}

\maketitle              
\begin{abstract}
Embedding a face image to a descriptor vector using a deep CNN is a widely used technique in face recognition. Via several possible training strategies, such embeddings are supposed to capture only identity information. Information about the environment (such as background and lighting) or changeable aspects of the face (such as pose, expression, presence of glasses, hat etc.) should be discarded since they are not useful for recognition. In this paper, we present a surprising result that this is not the case. We show that non-ID attributes, as well as landmark positions and the image histogram can be recovered from the ID embedding of state-of-the-art face embedding networks (VGGFace2 and ArcFace). In fact, these non-ID attributes can be predicted from ID embeddings with similar accuracy to a prediction from the original image. Going further, we present an optimisation strategy that uses a generative model (specifically StyleGAN2 for faces) to recover images from an ID embedding. We show photorealistic inversion from ID embedding to face image in which not only is the ID realistically reconstructed but the pose, lighting and background/apparel to some extent as well.
\end{abstract}
\section{Introduction}


State-of-the-art face recognition relies on the use of deep neural networks (usually CNNs) to embed a face image to an identity vector \cite{cao2018vggface2,schroff2015facenet,parkhi2015deep}. A measure of distance in this embedding space is used to represent dissimilarity in identity. The goal of training such networks is to minimise the within-class scatter while maximising the between-class scatter for all identities. The former goal necessitates that the embedding should depend only on the identity of the person in the image. Environmental conditions such as the lighting, background and properties of the camera as well as changeable aspects of the face such as pose, expression and the presence of accessories should not affect this embedding (i.e.~should not introduce within-class scatter). In other words, the embedding network should learn invariance to these factors.
In this paper, we explore a number of unanswered questions about some of the properties of these embeddings.

\para{ID to image inversion}
First, we ask whether it is possible to create an image from an ID vector that correctly recreates the identity of the person. Conventional text-based passwords are usually passed through a cryptographic hash function for storage \cite{al2011cryptographic}. Since these are pseudo one-way functions, it is extremely difficult to invert an encrypted password to cleartext, with brute force or dictionary attacks the only options. For password verification, only the encrypted version need be stored and a leak is not critical due to the difficulty of inversion.
It is tempting to assume that ID embeddings from face images possess similar characteristics. For example, the Face ID facial recognition system developed by Apple Inc.~makes this argument in its advertising to reassure users:
\begin{displayquote}
\emph{Face ID doesn’t store an image of your face. Instead of storing an image, Face ID saves a mathematical value created from the characteristics of your features. It’s impossible for anyone else to recreate your likeness from this.} \cite{faceID}
\end{displayquote}
However, since the embedding of a face image to an ID vector is a noisy process, it cannot be assumed that an identical embedding will arise from any image of the same person's face. Therefore, the ID vector cannot be passed through a hash function for storage since the hashed value will change dramatically with small changes in the ID vector. From a security perspective, this means that raw ID vectors must be stored for future identification. From an inversion perspective, it means that similar images map to similar ID vectors. Also, embeddings based on deep neural networks are computed via a differentiable function. This means that it is possible to optimise an image representation to minimise an ID vector loss, thereby reconstructing an image with the desired identity (subject to suitable regularisation, which we achieve via a generative model).

\para{Non-ID information leakage}
Second, we ask whether ID embeddings truly contain only ID-related information.
The engineering of training datasets, network architectures and loss functions has been widely studied in the face recognition literature in order to satisfy the goal of invariance to non-ID factors in the input image. Datasets are created specifically to introduce lots of variation in these non-ID factors. Then, by designing a loss function that encourages the same ID to embed to the same point, invariance to these factors is hopefully learnt. In this paper, we investigate to what extent this has been achieved. In particular, we ask: how well do modern face embedding networks successfully remove non-ID factors when embedding a face image? 

\para{Inversion including non-ID factors}
Finally, we connect these two lines of investigation by asking whether it is possible to recover an image from an ID vector that not only captures the correct face identity but also non-ID characteristics of the actual image that was used to compute the ID vector. The possibility of ID vector to image inversion raises privacy and security questions. 

For the reasons mentioned above, ID descriptor vectors cannot be stored securely hashed. This means that any third party with whom identity must be verified is receiving an encoding of a face image from which an image can be recovered. Where non-ID information leaks into this representation, it means the image itself can potentially be recovered. This could, for example, leak unintended information in the background of the image or maybe an unflattering image that the user would not wish to be made public.

\section{Related work}
\para{GAN inversion}
General adversarial networks (GANs) have become a popular tool in computer vision and machine learning. In 2019 and 2020, Karras et al.~\cite{DBLP:conf/cvpr/KarrasLA19,DBLP:conf/cvpr/KarrasLAHLA20} set the benchmark with StyleGAN and StyleGAN2, able to create photo-realistic face images at high resolution. A defining feature of StyleGAN is in its generator architecture. Instead of feeding an input latent code $ z \in Z $ only to the beginning of the network, a mapping network first transforms it to an intermediate latent code $ w \in W$. Affine transforms then produce styles that control the layers of a synthesis network. In addition to that, stochastic variation is facilitated by providing random noise maps to the synthesis network.

Recent work has shown that GANs can encode a rich set of semantics in their latent space. In addition to generating images, recently, attempts have been made to invert the GAN generating process from the image back to latent space for the purpose of image manipulation or analysis, which is widely known as GAN inversion. To accomplish that, most existing works either learn an extra encoder or regressor separate from the GAN (e.g.~\cite{bayat2020inverse,richardson2020encodingStylGAN}), or directly optimise the latent code to fit a target image (e.g.~\cite{DBLP:journals/corr/abs-2104-07661}), or a combination of the two (i.e.~initialising the optimisation with the result from a regressor, e.g.~\cite{zhu2020domaininversion}).

For StyleGAN specifically, recently, several works have shown that it is possible to retrieve the latent code $ w $ of a target image~\cite{shen2020interpreting,tewari2020stylerig}. The works show that inverting to the latent space $W$ is easier than to $Z$. However, accurately reconstructing a target image is still an ongoing challenge. In another recent work, Abdal et al.~\cite{abdal2019image2stylegan,abdal2020image2stylegan++} propose a framework to project an image into the latent space $W+$, where $W+$ contains separates latent vectors for the specific scales of StyleGAN, hence effectively reconstructing different levels of features in the target image.
Finally, Yin et al.~\cite{yin2020dreaming} introduce a deep-inversion method that inverts a pretrained neural network to generate synthesised class-conditional input images for data-free knowledge transfer.

What all these methods have in common is that the inversion they do usually means going from an image to a latent code.
Much different to that, we investigate \emph{ID-descriptor to image} inversion, which uses StyleGAN as a generative model (i.e. ID$\rightarrow$StyleGAN code$\rightarrow$image).

\para{ID descriptor information \& inversion}
In a seminal work, in 2015, Mahendran and Vedaldi~\cite{DBLP:conf/cvpr/MahendranV15} set out to analyse the visual information contained in both shallow (e.g.~HOG) and deep feature representations, to investigate the question: \emph{given an encoding of an image, to which extent is it possible to reconstruct the image itself}. They propose an optimisation method to invert representations using gradient descent. Among their findings are that networks retain rich information even at deep levels and that a progressively more invariant and abstract notion of the image content is formed in the network. As face identity descriptors usually are made up of the final layer of a deep network, it would therefore be the most invariant and abstract representation.

Few works so far have investigated the inversion of a face descriptor back to a face image. Genova et al.~\cite{genova2018unsupervised} train an image to 3DMM parameter regressor in an unsupervised manner, where the key idea is an ID loss between ID descriptor of the original image and the ID descriptor of a 3DMM image rendered with a differentiable renderer. The only trainable part of their system is the \emph{ID descriptor to 3DMM parameter} regressor. This constitutes an ID inversion network - but they do not reconstruct the image itself, so no non-ID information is recovered.

Following on from the general idea of an ID loss, there are a number of works that employ an identity loss in 3D face model fitting, for example GANFIT~\cite{DBLP:conf/cvpr/GecerPKZ19}.
Cole et al.~\cite{DBLP:conf/cvpr/ColeBKSMF17} also perform an ID-only inversion for the purposes of face frontalisation. They assume that the face encoder successfully removes all non-ID information and train to reconstruct only frontal image landmarks and textures. Since they do not use a GAN, their results are not photorealistic.
Some recent works~\cite{DBLP:conf/eccv/RazzhigaevKKTP20,duong2020vec2face} have investigated a so-called \emph{black-box attack}, which is whether given only an ID descriptor, and no access to an attacked model, one could reconstruct an image of the face, and they have presented encouraging (technically) as well as concerning (from a privacy perspective) results.

In terms of non-ID information contained in face descriptors, in an early work, Kumar et al.~\cite{DBLP:conf/iccv/KumarBBN09} came to the perhaps surprising result that using a so-called \emph{inverse crop}, where the face is cut out of an image, leads to surprisingly high face recognition rates on LFW. It should be noted though that these inverse crops do contain hair and part of ears/chin, and that LFW is a fairly simple dataset (recognition rates achieved are over 99\%, even a decade ago) - so it is largely unexplored, if any background or other non-ID information is present in face descriptors, especially in today's state-of-the-art networks.
To the best of our knowledge, no work so far has investigated if any non-ID properties can be recovered from identity descriptors.


\para{Privacy leakage \& Adversarial learning}
Past studies have shown that face recognition networks encode soft biometrics (e.g.~age, race and gender) while training~\cite{fu2010age}. This indicates that face descriptors are at risk of privacy leaks. Privacy leakage in face representation is a vital issue, as it can raise several concerns associated with the unauthorised extraction of an individual’s information. More recent studies have shown that the extraction of such soft biometric data can improve the performance of facial recognition systems~\cite{mirjalili2018semiImpartingPrivacyFaceImages,dhar2020genderneutral}. 

Inspired by GANs, several adversarial learning methods have been proposed that remove or mitigate sensitive information stored in the training data or learn somehow obfuscated features.
Alvi et al.~\cite{alvi2018turning} are inspired by domain and task adaptation methods, they propose a joint learning and unlearning method to remove bias from neural network embedding. To unlearn the bias, the authors adversarially minimise the classification loss and confusion loss. The confusion loss is computed by calculating the cross-entropy between classifier output and uniform distribution.
Li et al.~\cite{li2019deepobfuscator} deployed several neural networks to simulate an attacker that attempts to reconstruct entire raw data. An obfuscator is trained to hide privacy-related information, which ensures that the attacker cannot train using features provided by the obfuscator to accurately infer privacy attributes or reconstruct the raw data.
Dhar et al.~\cite{dhar2020genderneutral} propose to adversarially minimise gender predictability to reduce gender bias from pretrained face descriptors. They show that gender bias in face recognition is correlated with the power of face descriptors to predict gender. The face descriptors with low gender predictability generally demonstrated lower gender bias in face verification.


\section{Non-ID attribute prediction from ID}

\begin{figure}[!t]
    \centering
    \includegraphics[width=0.7\columnwidth,trim={55 50 70 55},clip]{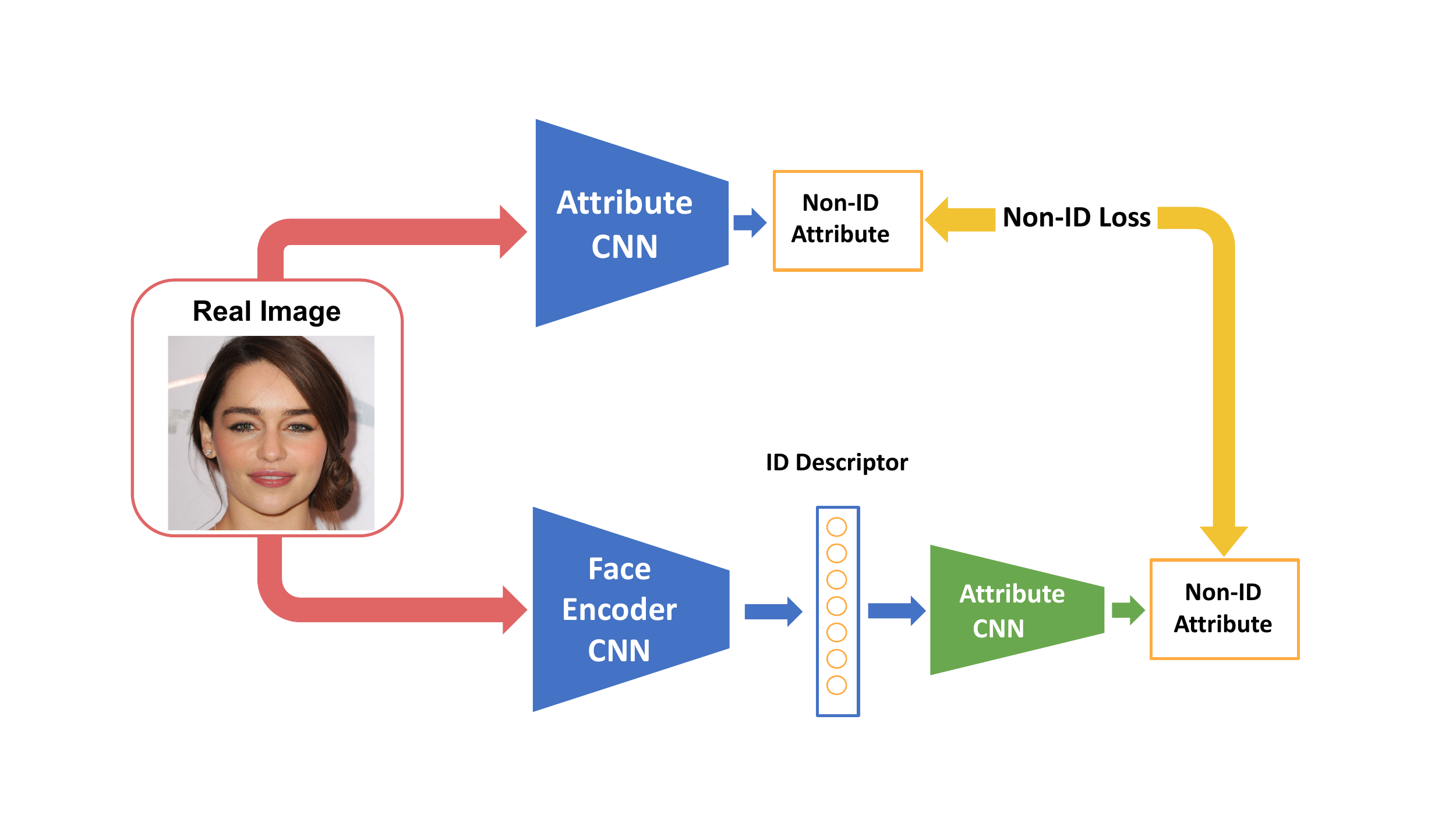} 
    \caption{Non-ID attribute regression via an ID bottleneck. Only the green component is trained: an MLP that maps an ID vector to the appropriate attribute (such as expression, landmarks, image histogram etc). The labels either come from a pretrained (and fixed) attribute estimation network that takes an image as input, they are provided as manually assigned labels or they are computed directly from the  image.}
    \label{fig:non_id_attribute_regression}
\end{figure}

We begin by exploring to what extent we are able to estimate non-ID ``attributes'' from an ID descriptor provided by a pretrained face encoder CNN. We use ``attribute'' here in very general terms, including image-based attributes such as landmark positions and colour histograms and non-ID face attributes such as the presence or absence of a smile, glasses or hat. For each attribute, we train an MLP that maps from an ID descriptor to the target attribute. All of our MLPs are trained on CelebA \cite{liu2015celeba}, which includes 202,599 celebrity face images with 40 binary attributes.  We aligned and cropped the original CelebA to a VGGFace2 compatible version and scaled all images to resolution $224$.
We train in a supervised fashion using either labelled real data or synthetically generated data. In all cases, we embed images $\mathbf{i}$ to an ID descriptor,
$\mathbf{d}=\text{VGG}(\mathbf{i})$ using the VGGface2 face encoder \cite{cao2018vggface2}, where $\mathbf{d}\in\mathbb{R}^{2048}$. Fig.~\ref{fig:non_id_attribute_regression} illustrates our proposed Non-ID attribute regression framework. 

\subsection{Discrete Binary Attributes}

We begin by estimating discrete binary attributes. These have been manually labelled as part of the CelebA \cite{liu2015celeba} dataset. Then we train an MLP to classify the binary class using Binary Cross Entropy loss, Adam with learning  rate $10^{-3}$, for $20$ epochs. We train separate networks for the smiling, glasses and wearing hat attributes. Our MLP consists of $3$ fully connected layers with $256$ hidden neurons and a sigmoid output layer.

\begin{table}[!t]
\centering
\begin{tabular}{@{}l|c|c|c|c|c@{}}
           & \multicolumn{3}{c|}{Attribute}      & Landmarks & Histogram \\
           & Smiling & Wearing\_Hat & Eyeglasses & mean      & EMD     \\
From ID (VGGFace2 \cite{cao2018vggface2})    & 91.0\%       & 99.0\%            & 99.7\%          & 9.3\%         & 2.68       \\
From ID (ArcFace \cite{deng2019arcface})    & 81.7\%       & 96.7\%            & 96.7\%          & 7.3\%         & 2.45       \\
From image \cite{liu2015celeba} & 92\%     & 99\%          & 99\%        & 8.2\%         & 0.0        \\
Baseline      & 50.4\%       & 96.7\%            & 94.0\%          & 11\%         & 2.97   
\end{tabular}%
\vspace{0.4em} 
\caption{Quantitative results for attribute prediction (discrete binary attributes, landmarks and image histogram) from ID vectors (row 1 and 2) and images (row 3). In row 4 we show baseline performance in which we simply always predict the most common class, the mean landmarks or the mean histogram respectively. 
For attribute prediction, higher is better, and for the landmark and histogram prediction, lower is better.
}
\label{tab:4.3result}
\end{table}

\subsection{Histogram regression}\label{sec:histreg}
In this section, we examine whether background colour and lighting information can be restored from ID by predicting the histogram of the RGB channel. Histograms of RGB intensities provide a global summary of an image that encapsulates not only ID but also environment-related features such as background, camera settings and lighting. For each image, we compute a ground truth hard histogram and use this as the label for training. Here, the three colour values are binned into fixed width bins, with one histogram per colour channel. We found that a very small MLP provides best performance for the task of image histogram regression from ID vector. We use $2$ fully connected layers with ReLU activation and $8$ neurons per layer. We apply softmax to the output layer such that the output represents a normalised histogram. We use $N=10$ bins. We train this network using mean squared error loss, Adam with learning rate of $10^{-6}$, and batch size $32$, for $20$ epochs.

\subsection{Landmark regression}\label{sec:landmarkreg}

Finally, we attempt to directly regress the coordinates of 68 face landmarks from the ID vector. We apply the dlib landmark detector \cite{kazemi2014onedlib} to all images in the CelebA dataset \cite{liu2015celeba}. We use these as pseudo ground truth labels for our regressor. We use a three-layer MLP to predict landmarks from ID vector with 256 neurons per hidden layer and 136 outputs for the 2D coordinates of each landmark. For comparison, we also train a more conventional image to landmark regressor using a CNN. We use a simple architecture comprising 5 convolutional layers followed by 2 fully connected layers. The activation function is ReLUs. 
Both image to landmark and ID vector to landmark networks are trained using mean squared error loss and the SGD optimiser with learning rate $10^{-3}$, batch size $16$, and for $150$ epochs.
\subsection{Results}


\begin{figure}[!t]
    \centering
    \resizebox{0.9\textwidth}{!}{
    \begin{tabular}{c@{\hspace{0.1cm}}c@{\hspace{0.1cm}}c@{\hspace{0.1cm}}c@{\hspace{0.1cm}}c@{\hspace{0.1cm}}|c@{\hspace{0.1cm}}c@{\hspace{0.1cm}}c}
    & \multicolumn{4}{c|}{\Large Correctly classified} & & \multicolumn{2}{c}{\Large Incorrectly classified} \\
    \rotatebox[origin=l]{90}{\hspace{0.25cm}{\Large Smiling}} & \includegraphics[height=0.3\columnwidth]{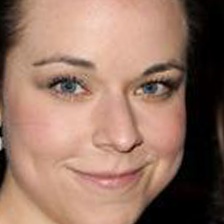} &
    \includegraphics[height=0.3\columnwidth]{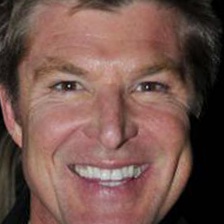} &
    \includegraphics[height=0.3\columnwidth]{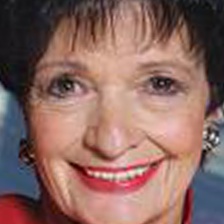} &
    \includegraphics[height=0.3\columnwidth]{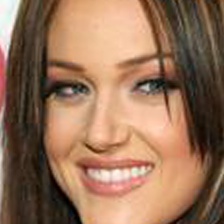} &
    \rotatebox[origin=l]{90}{{\Large False positives}} &
    \includegraphics[height=0.3\columnwidth]{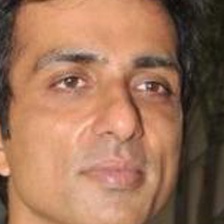} &
    \includegraphics[height=0.3\columnwidth]{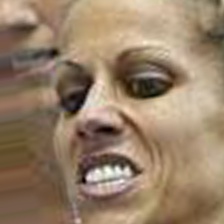} \\
    \rotatebox[origin=l]{90}{{\Large Wearing hat}} &
    \includegraphics[height=0.3\columnwidth]{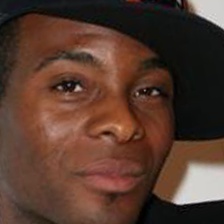} &
    \includegraphics[height=0.3\columnwidth]{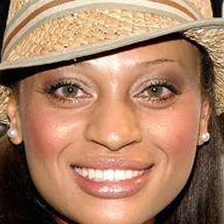} &
    \includegraphics[height=0.3\columnwidth]{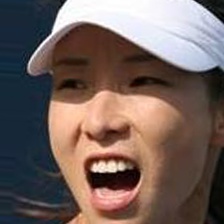} &
    \includegraphics[height=0.3\columnwidth]{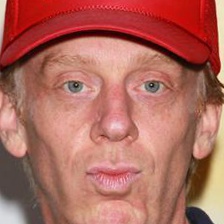} &
    \rotatebox[origin=l]{90}{{\Large False positives}} &
    \includegraphics[height=0.3\columnwidth]{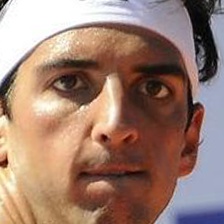} &
    \includegraphics[height=0.3\columnwidth]{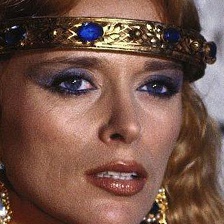} \\
    \rotatebox[origin=l]{90}{{\Large Eyeglasses}} &
    \includegraphics[height=0.3\columnwidth]{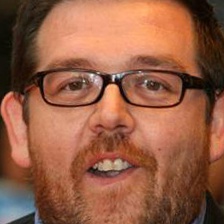} &
    \includegraphics[height=0.3\columnwidth]{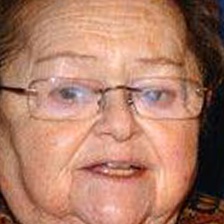} &
    \includegraphics[height=0.3\columnwidth]{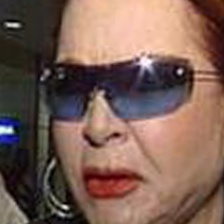} &
    \includegraphics[height=0.3\columnwidth]{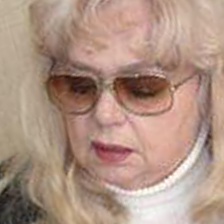} &
    \rotatebox[origin=l]{90}{{\Large False negatives}} &
    \includegraphics[height=0.3\columnwidth]{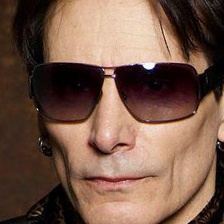} &
    \includegraphics[height=0.3\columnwidth]{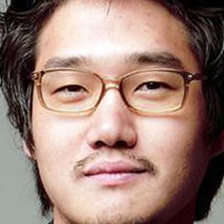} \\
    \end{tabular}
    }
    \caption{Examples of correctly (left) and incorrectly (right) classified samples. We show the original images but note that the classification is done \emph{only on the ID vectors derived from these images}.}
    \label{fig:qualattpred}
\end{figure}

We now evaluate our prediction of non-ID attributes from ID vectors, as provided by the VGGFace2 network. We show quantitative results for all attributes in Table~\ref{tab:4.3result}. The evaluation images we used are 15k test images from CelebA with the remaining 188k images for training. 
To validate our results, we repeat the experiments with ID vectors generated by the ArcFace network.

For discrete binary attributes, we show the percentage classified correctly. Our result regressed from the ID vector is shown in the first and second rows. This shows that non-ID leakage exists in different networks. For comparison in the third row we show the result from \cite{liu2015celeba} computed \emph{from the original image}. In the fourth row, we show the baseline performance obtained by always guessing the more common class for the binary attribute prediction, the mean landmarks for the landmark prediction, and the mean histogram for the histogram prediction. With the exception of the wearing hat attribute for the ArcFace embedding, we can see that we significantly outperform that baseline and, remarkably, match or even exceed the performance of an image-based method despite only having access to an ID vector that should be independent of these non-ID attributes. In Fig.~\ref{fig:qualattpred} we show some examples of correctly and incorrectly classified samples. It is interesting to note that quite subtle smiles are encoded in the ID vectors such that we correctly classify them and, even in the case of the false positives shown, there are still smile-like features in the wrinkles around the mouth. Similarly, the false positives for wearing hat are in fact wearing headgear.

\begin{figure}[!t]
    \centering
    \resizebox{\textwidth}{!}{
    \begin{tabular}{@{}c@{\hspace{0.01cm}}c@{\hspace{0.01cm}}c@{\hspace{0.01cm}}c@{\hspace{0.01cm}}c@{\hspace{0.01cm}}c@{}}
        \includegraphics[height=0.5\columnwidth]{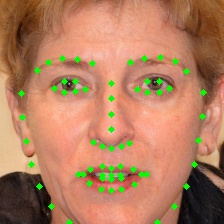} &
        \includegraphics[height=0.5\columnwidth]{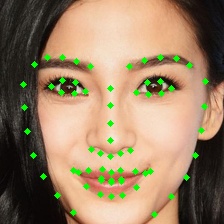} &
        \includegraphics[height=0.5\columnwidth]{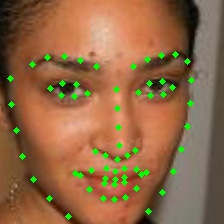} &
        \includegraphics[height=0.5\columnwidth]{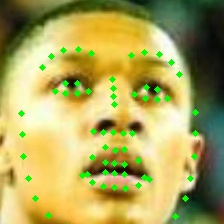} &
        \includegraphics[height=0.5\columnwidth]{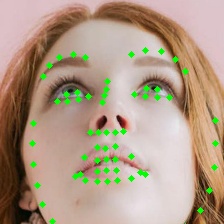} \\
        \includegraphics[height=0.5\columnwidth]{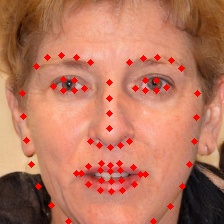} &
        \includegraphics[height=0.5\columnwidth]{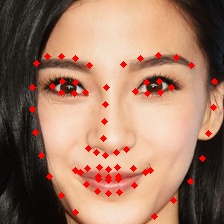} &
        \includegraphics[height=0.5\columnwidth]{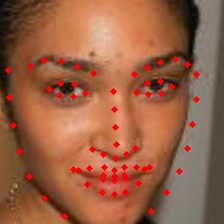} &
        \includegraphics[height=0.5\columnwidth]{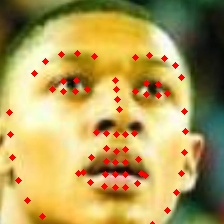} &
        \includegraphics[height=0.5\columnwidth]{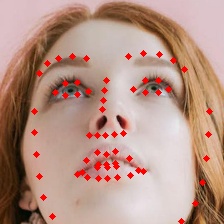} \\
        \includegraphics[height=0.5\columnwidth]{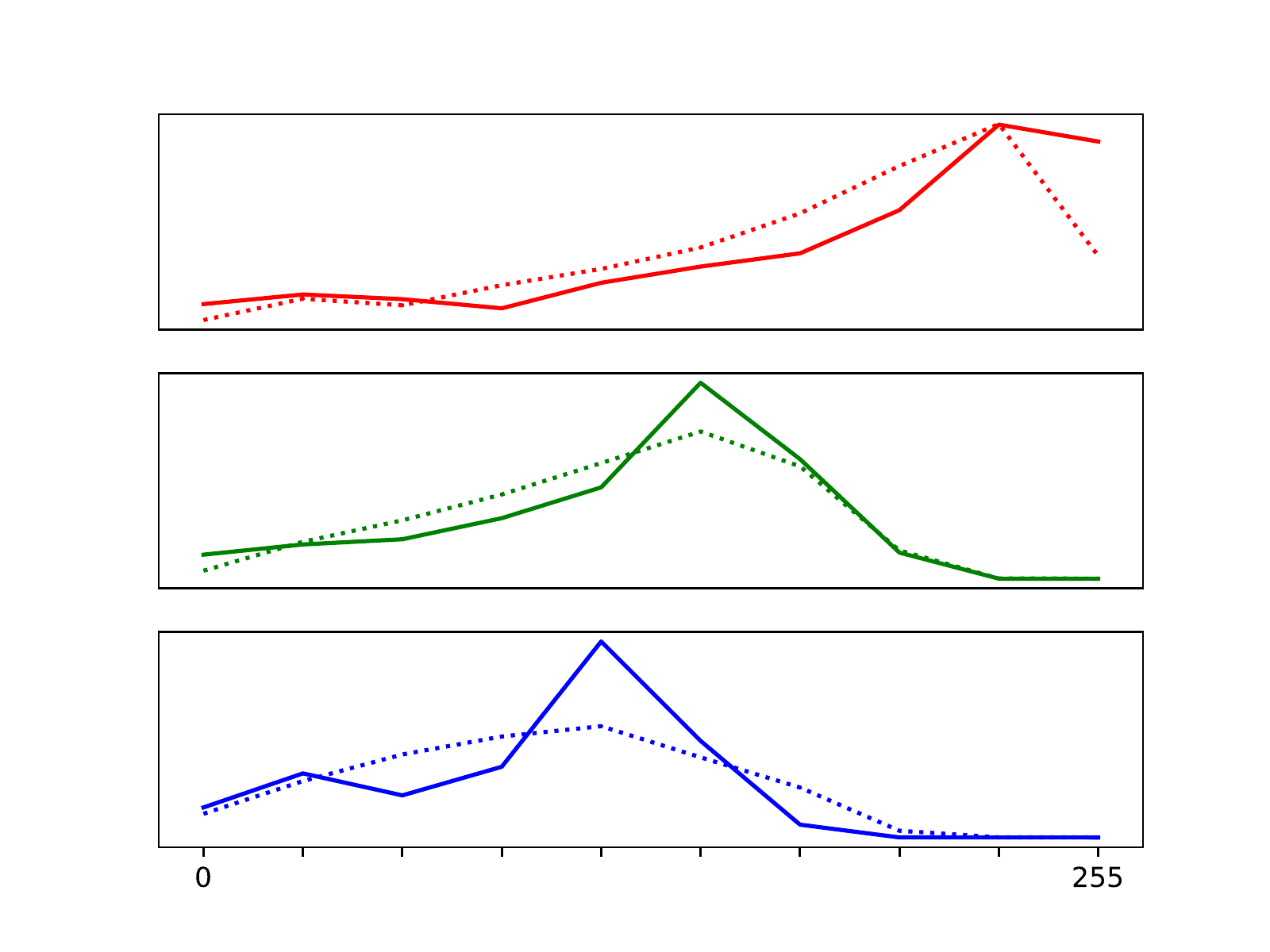} &
        \includegraphics[height=0.5\columnwidth]{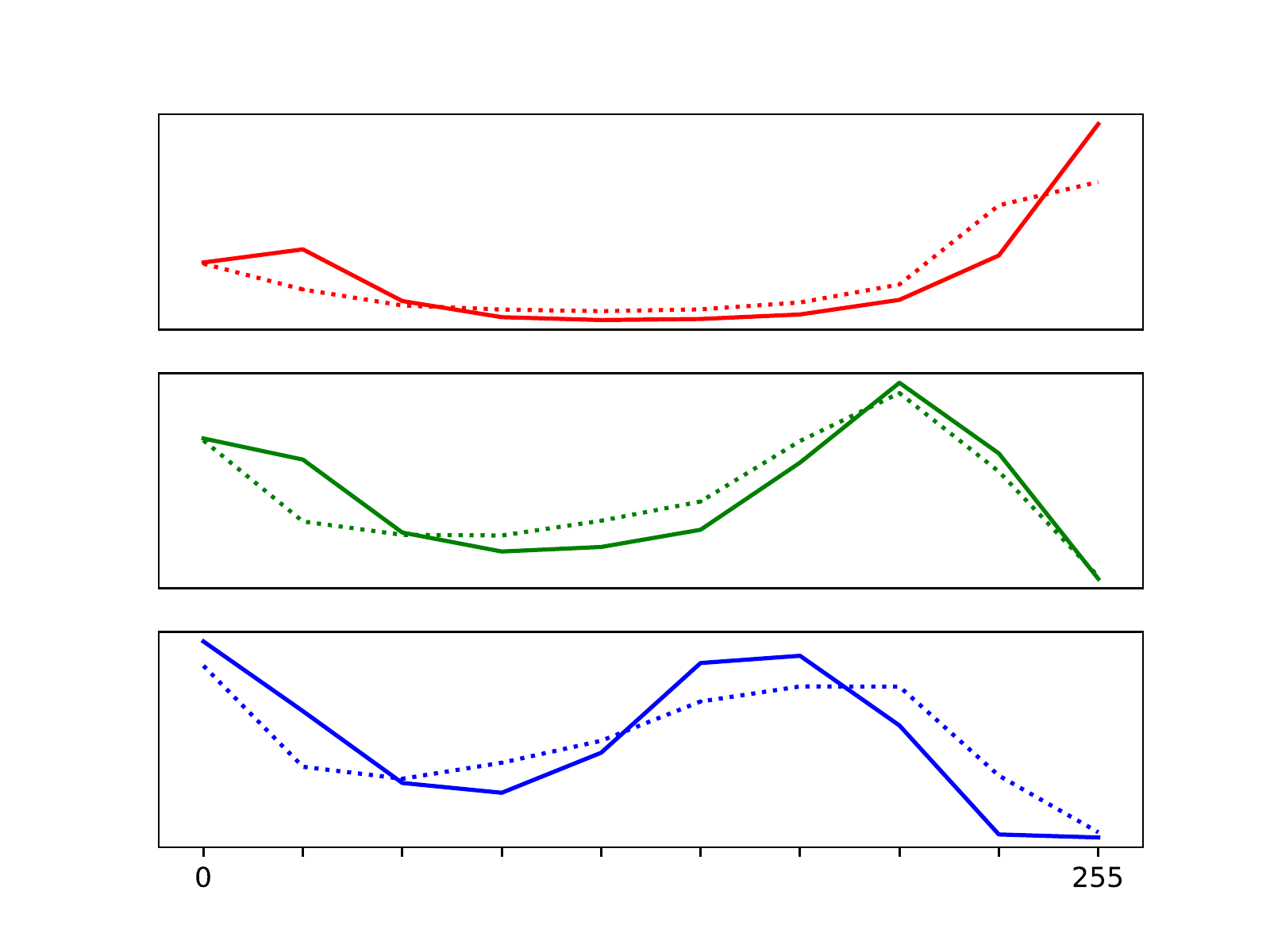} &
        \includegraphics[height=0.5\columnwidth]{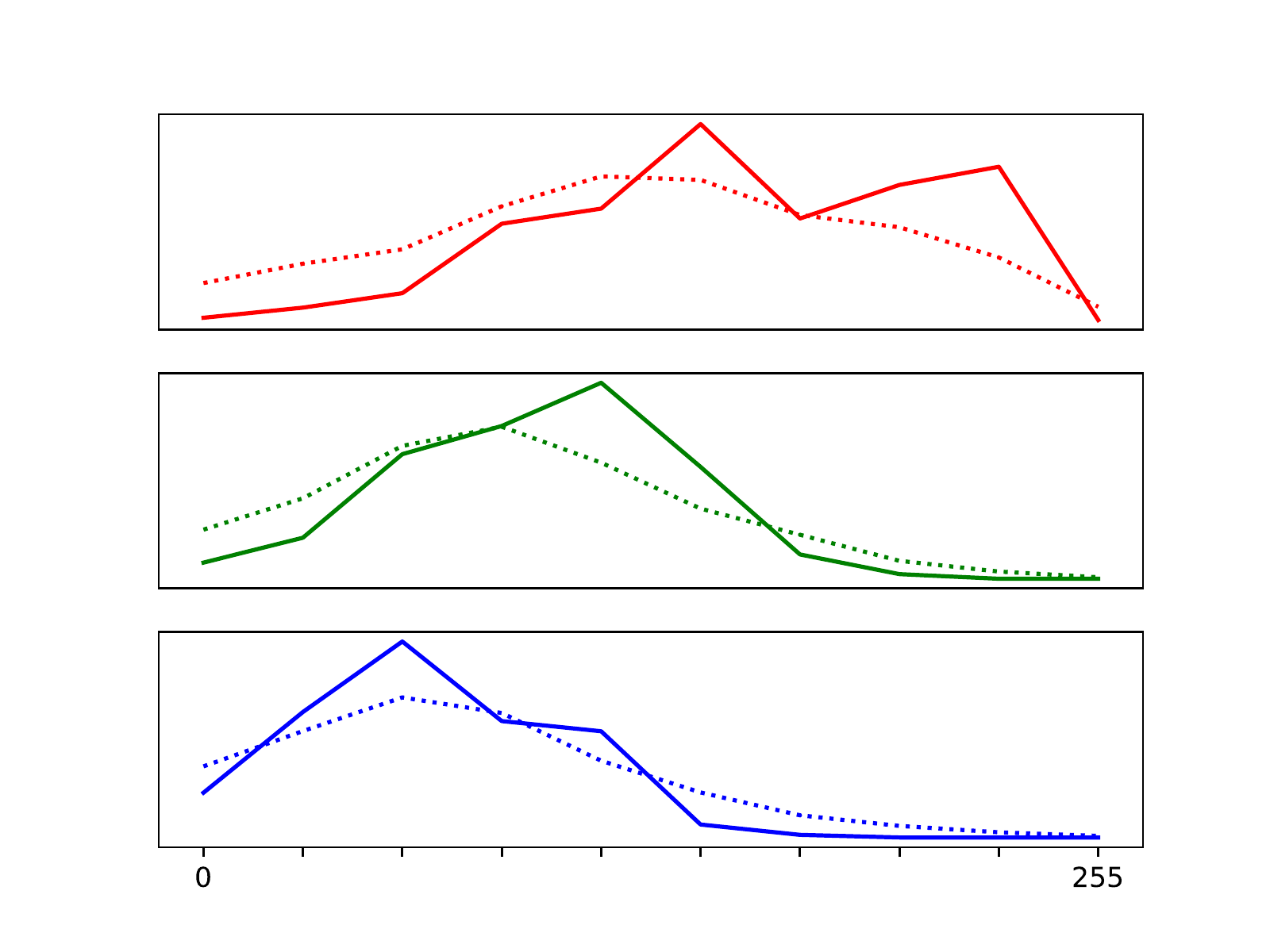} &
        \includegraphics[height=0.5\columnwidth]{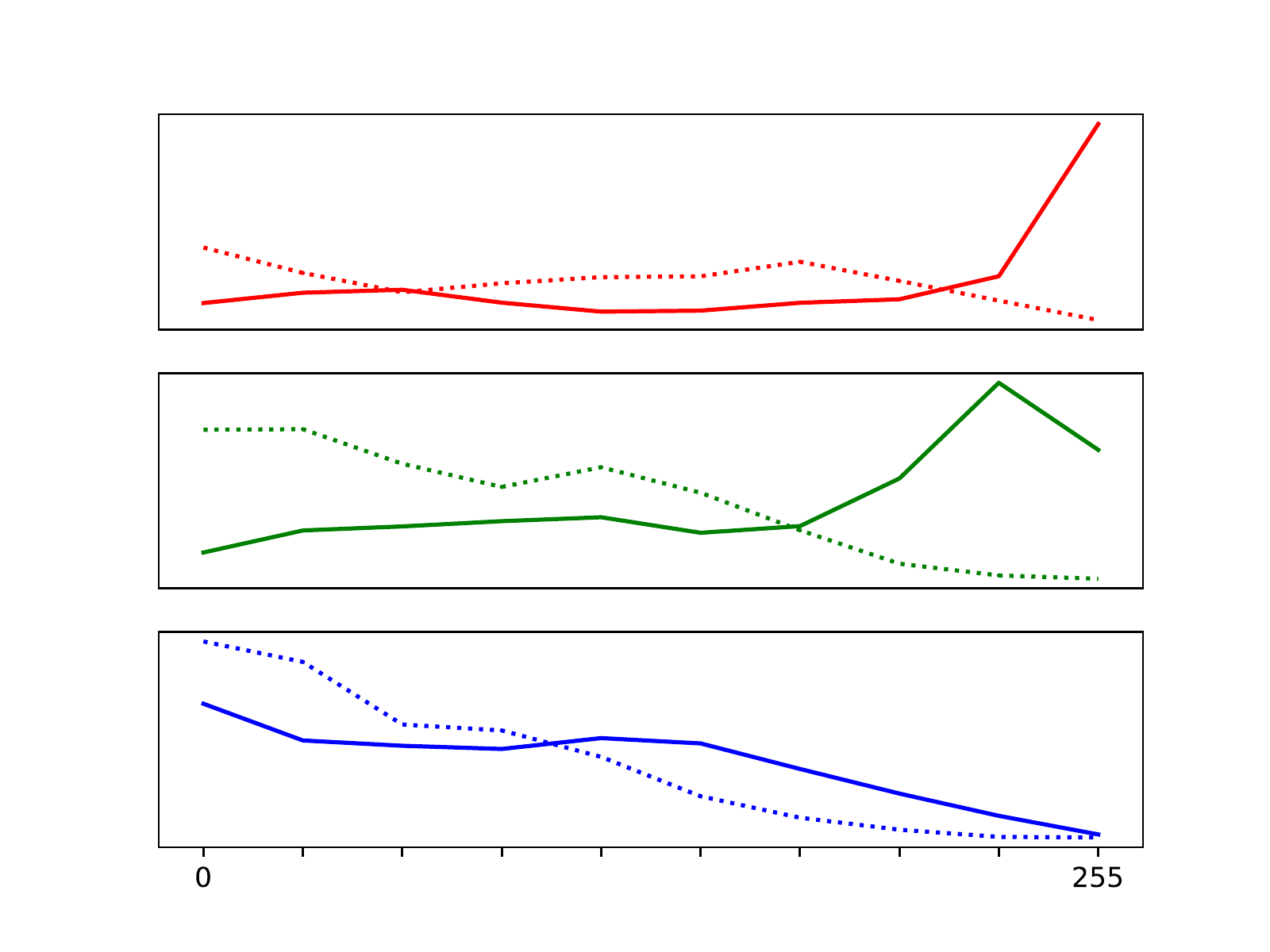} &
        \includegraphics[height=0.5\columnwidth]{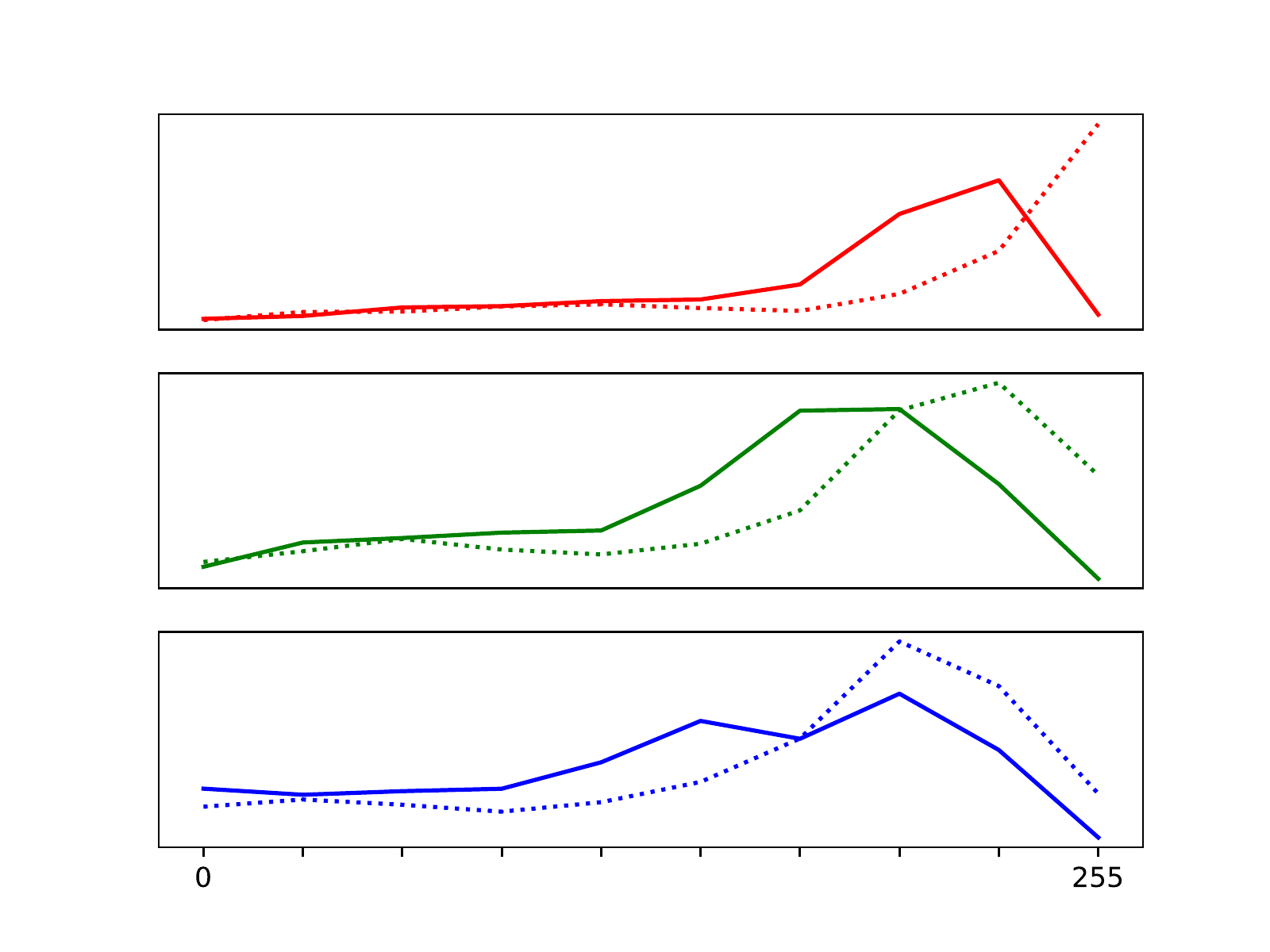} \\
    \end{tabular}
    }
    \caption{Qualitative results for histogram and landmark regression. row 1: input image with ground truth landmarks, row 2: landmarks regressed from ID vector, row 3: ground truth image histograms (dotted) and histograms regressed from ID vector (solid).}
    \label{fig:qualhistland}
\end{figure}

We now evaluate image-based attributes. In Table~\ref{tab:4.3result} we show quantitative results for landmark prediction in the fifth column and histogram prediction in the sixth column. For the landmark error, we show Euclidean distance averaged over landmarks expressed as a percentage of the interocular distance. For histogram error, we show the Earth Mover's distance to ground truth. Our prediction from ID vector outperforms the baseline and is only marginally worse than prediction from images (in the case of landmarks). In the case of histogram, the prediction from images is exact. In Fig.~\ref{fig:qualhistland} we show qualitative results for landmark and histogram estimation from ID vectors. In the first row we show the original image with ground truth (dlib) landmarks overlaid. In the second row we show the original images with the landmarks regressed from the ID vector overlaid. The landmarks are qualitatively convincing and clearly reconstruct pose - an entirely non-ID related property. In the third row we show the ground truth (dotted lines) and estimated (solid lines) RGB image histograms.

\section{Image from ID with a generative model}

We now investigate if, and how well, an image of a person can be recovered from an ID descriptor. Having shown that non-ID attributes can be estimated from a face descriptor, we also explore to what extent the original image itself, including non-ID information, can be reconstructed from an ID descriptor.

\begin{figure}[!t]
    \centering
    \includegraphics[width=\textwidth]{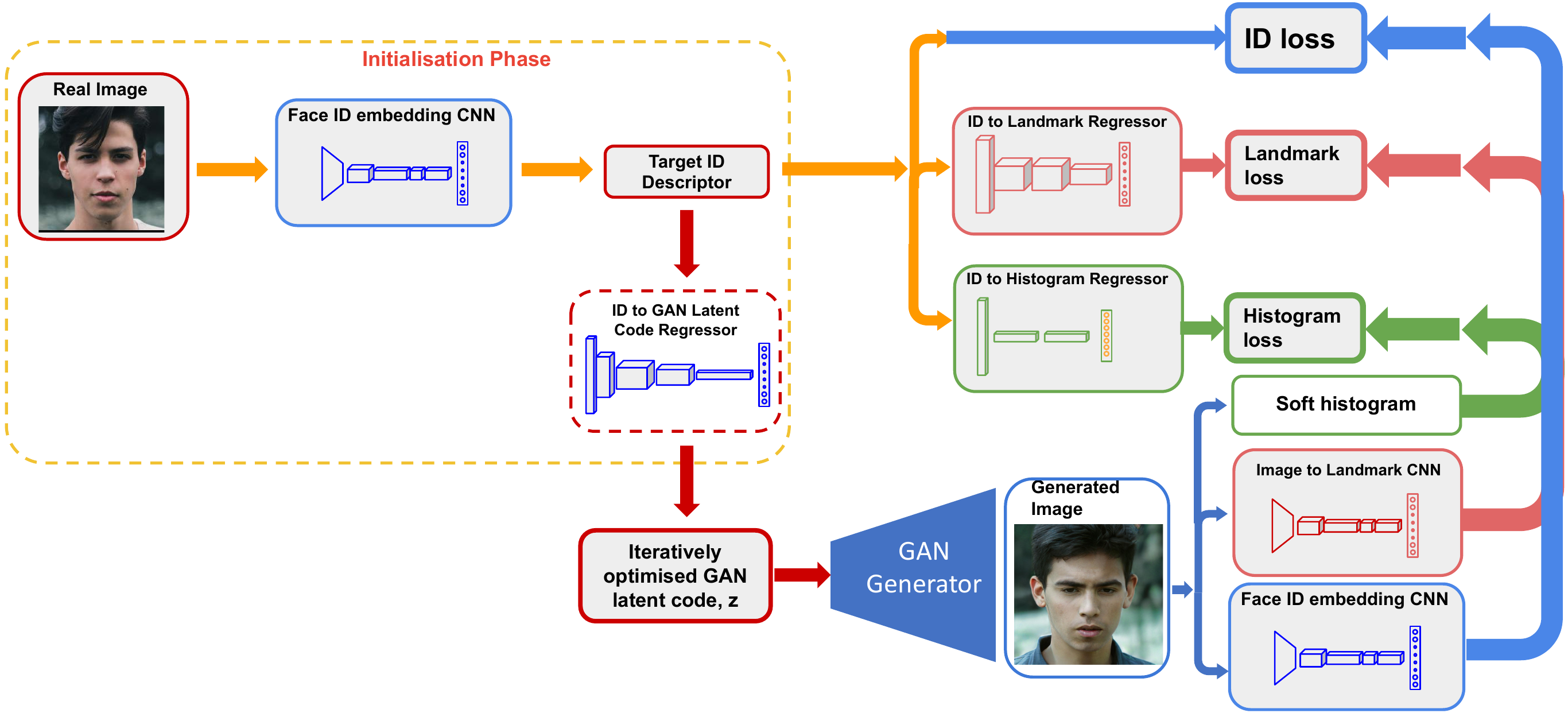}
    \caption{Reconstructing an image from an ID descriptor, including preservation of non-ID properties (landmarks and image histogram). We assume we only have access to the ID descriptor of a real image. We initialise the optimisation using a regression network to predict GAN latent code from ID descriptor. We then iteratively optimise the GAN latent code in order to produce an image that matches the ID, landmarks and histogram predicted from the target ID descriptor using pretrained networks.}
    \label{fig:descriptor_inversion_with_additional_losses}
\end{figure}

\subsection{ID-only inversion}\label{sec:IDonlyinv}

We begin by attempting to create an image that is recognisably the same person as the original but not necessarily similar to the original image. We pose this as an optimisation problem and use a generative face model to constrain the problem. Specifically, we restrict the solution to the space of images represented by the StyleGAN2 face model~\cite{DBLP:conf/cvpr/KarrasLAHLA20}. We denote by $\mathbf{i}=g(\mathbf{z})$ the face image that arises via the generator from the latent code $\mathbf{z}$. Suppose we are given a target VGG descriptor, $\mathbf{d}$, then we wish to solve the following optimisation problem:
\begin{equation}
    \min_{\mathbf{z}} L_{\text{ID}}(\mathbf{z}),\quad \text{where }
    L_{\text{ID}}(\mathbf{z}) = \left\| \text{VGG}(g(\mathbf{z})) - \mathbf{d} \right\|^2_2. \label{eqn:IDinv}
\end{equation}
In practice, this optimisation is prone to convergence on local minima and sensitive to initialisation. For this reason, we train a network that we use for initialisation that regresses a StyleGAN2 latent code directly from an ID descriptor. Our network comprises an MLP with 3 hidden layers with ReLU activation, 2,048 units per hidden layer. We train this network using synthetic data obtained by randomly sampling images from StyleGAN2, passing these through the face encoder network and then using the resulting ID descriptor and random GAN latent code as an input/output training pair. We subsequently optimise \eqref{eqn:IDinv} using the Adam optimiser with a learning rate $0.001$.

\begin{figure}[!t]
    \centering
    \resizebox{0.9\textwidth}{!}{
    \begin{tabular}{@{}c@{\hspace{0.1cm}}c@{\hspace{0.03cm}}c@{\hspace{0.03cm}}c@{\hspace{0.03cm}}c@{\hspace{0.03cm}}c@{}}
        \rotatebox[origin=l]{90}{\hspace{0.7cm} Input} & 
        \includegraphics[height=0.25\columnwidth]{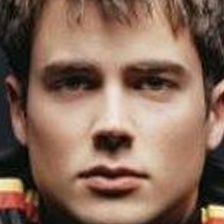} &
        \includegraphics[height=0.25\columnwidth]{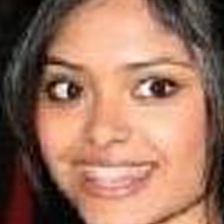} &
        \includegraphics[height=0.25\columnwidth]{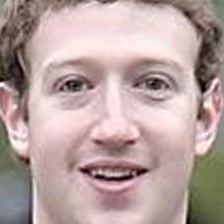} &
        \includegraphics[height=0.25\columnwidth]{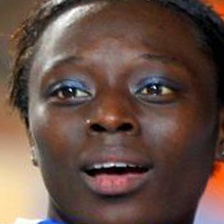} &
        \includegraphics[height=0.25\columnwidth]{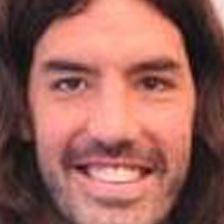} \\
        \rotatebox[origin=l]{90}{\hspace{-0.0cm} Direct regression} & 
        \includegraphics[height=0.25\columnwidth]{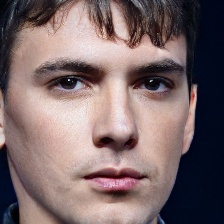} &
        \includegraphics[height=0.25\columnwidth]{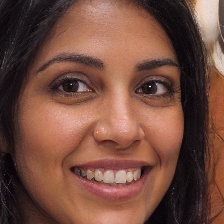} &
        \includegraphics[height=0.25\columnwidth]{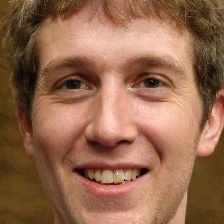} &
        \includegraphics[height=0.25\columnwidth]{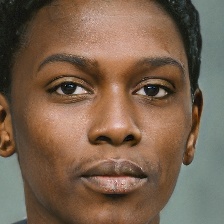} &
        \includegraphics[height=0.25\columnwidth]{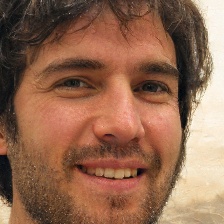} \\
        \rotatebox[origin=l]{90}{\hspace{0.1cm} ID optimisation} & 
        \includegraphics[height=0.25\columnwidth]{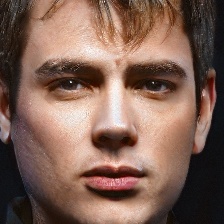} &
        \includegraphics[height=0.25\columnwidth]{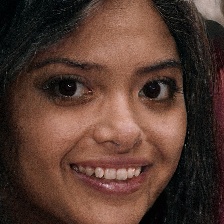} &
        \includegraphics[height=0.25\columnwidth]{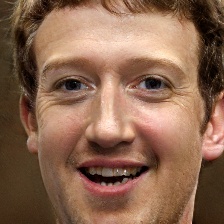} &
        \includegraphics[height=0.25\columnwidth]{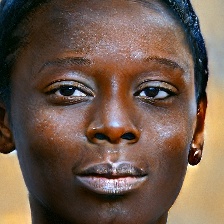} &
        \includegraphics[height=0.25\columnwidth]{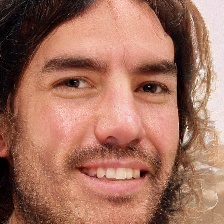} \\
    \end{tabular}
    }
    \caption{Direct regression versus ID loss optimisation. Top row: input images. Middle row: output of ID to StyleGAN2 latent code regression network. Bottom row: after subsequent optimisation of StyleGAN2 latent code to minimise $L_{\text{ID}}$.}
    \label{fig:regression_vs_optimising_id_loss}
\end{figure}

\subsection{Image reconstruction}

The results of the above process successfully produce an image with the correct identity. However, they often fail to reconstruct certain features of the original image, for example the pose and expression of the face, the lighting in the image, the background and the presence of apparel. We have shown that, with suitable supervision and training, it is possible to extract some of these properties from weak signals that find their way into the ID descriptor. Once reconstructed, we now show that these can be used to provide additional, direct supervision to the inversion problem. Essentially, we ask that not only the ID be reconstructed but that additional, non-ID, features estimated from the ID descriptor also be reconstructed (specifically landmarks and image histogram).

\para{Landmarks} From the target ID descriptor, $\mathbf{d}$, we use the pretrained regression network described in Section~\ref{sec:landmarkreg} to compute approximate target landmarks, $f_{\text{ID}\rightarrow \text{landmarks}}(\mathbf{d})$. During reconstruction, we compare the target landmarks with those extracted from the current image reconstruction using the pretrained image to landmark regression CNN, $f_{\text{image}\rightarrow \text{landmarks}}(\mathbf{i})$:
\begin{equation}
    L_{\text{landmarks}}(\mathbf{z}) = \left\| f_{\text{image}\rightarrow \text{landmarks}}(g(\mathbf{z})) - f_{\text{ID}\rightarrow \text{landmarks}}(\mathbf{d}) \right\|^2_2.
\end{equation}

\para{Soft histogram} For the histogram reconstruction loss, we follow a similar strategy. We use the pretrained regression network described in Section~\ref{sec:histreg} to compute an approximate target histogram, $f_{\text{ID}\rightarrow \text{histogram}}(\mathbf{d})$. The exact histogram of the reconstructed image is discrete and therefore not differentiable. For this reason, we use a differentiable soft approximation of the image histogram.

The idea is to use sigmoid to softly assign values to bins. Consider a vector $\mathbf{x}\in\mathbb{R}^M$ of $M$ values. We wish to compute a soft histogram $H(\mathbf{x})\in\mathbb{R}^N$ which softly assigns all values in $\mathbf{x}$ to $N \in \mathbb{Z}^+$ histogram bins. We specify minimum and maximum values (we use $\text{min}=0$ and $\text{max}=255$ for image histograms) and the bin width by $\delta = \frac{\max - \min}{N}$. The $i$th bin centre is given by $c_i=\text{min}+\delta(i-0.5)$. Then, the value of the $k$th bin in $H$ is:
\begin{equation}
    H(\mathbf{x})_k = \sum_{j=1}^M f(\mathbf{x}_j-c_k+\delta/2) - f(\mathbf{x}_j-c_k-\delta/2),\label{SoftHist}
\end{equation}
where $f$ is an assignment function. In a hard (non-differentiable) histogram, $f$ is the Heaviside step function. In our soft histogram, we use sigmoid, $f(x) = \mathbf{Sigmoid}(\sigma z)$, with parameter $\sigma$ which controls the softness of the bins. When $\sigma$ is very large, the soft histogram approaches the hard histogram but the gradient vanishes, while small $\sigma$ yields a very soft histogram that badly approximates the true histogram. We use $\sigma = 1.85$ in our experiments. To compute a soft image histogram, we apply \eqref{SoftHist} to all values in one colour channel of an image, yielding three histograms.
Now we can write the histogram loss as:
$
    L_{\text{histogram}}(\mathbf{z}) = \left\| H(g(\mathbf{z})) - f_{\text{ID}\rightarrow \text{histogram}}(\mathbf{d}) \right\|^2_2
$.

\begin{figure}[!t]
    \centering
    \resizebox{0.8\textwidth}{!}{
    \begin{tabular}{c@{\hspace{0.1cm}}c@{\hspace{0.1cm}}c@{\hspace{0.1cm}}c}
    \small{Target} & \small{ID} & \small{ID \texttt{+}} & \small{ID \texttt{+} Landmark} \vspace{-0.1cm} \\& & \small{Landmark}  \vspace{-0.1cm} & \small{\texttt{+} Histogram} \vspace{+0.1cm} \\
    \includegraphics[height=0.27\columnwidth]{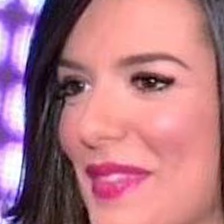} &
    \includegraphics[height=0.27\columnwidth]{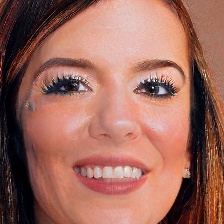} &
    \includegraphics[height=0.27\columnwidth]{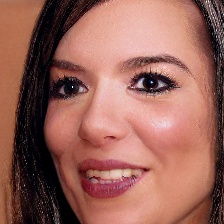} &
    \includegraphics[height=0.27\columnwidth]{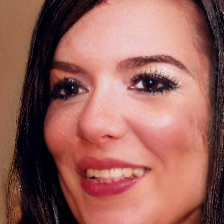} \\
    \includegraphics[height=0.27\columnwidth]{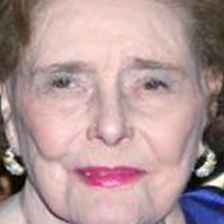} &
    \includegraphics[height=0.27\columnwidth]{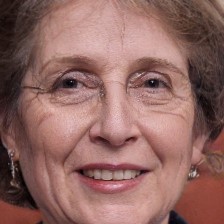} &
    \includegraphics[height=0.27\columnwidth]{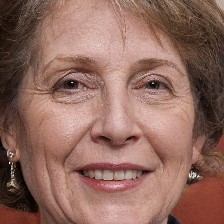} &
    \includegraphics[height=0.27\columnwidth]{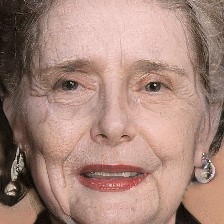} \\
    \includegraphics[height=0.27\columnwidth]{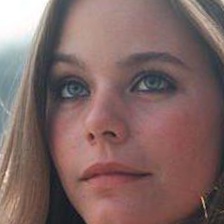} &
    \includegraphics[height=0.27\columnwidth]{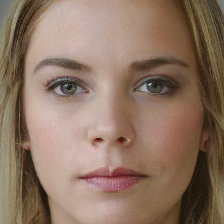} &
    \includegraphics[height=0.27\columnwidth]{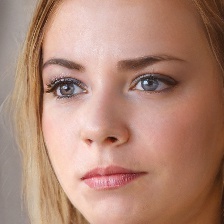} &
    \includegraphics[height=0.27\columnwidth]{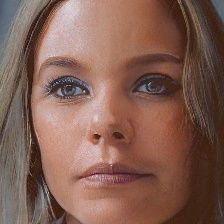}
    \\
    \includegraphics[height=0.27\columnwidth]{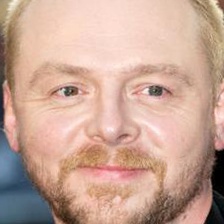}
    &\includegraphics[height=0.27\columnwidth]{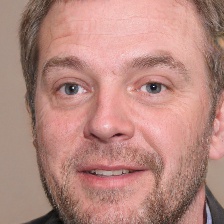}
    &\includegraphics[height=0.27\columnwidth]{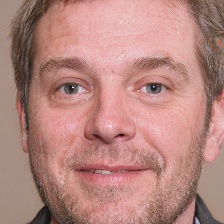}
    &
    \includegraphics[height=0.27\columnwidth]{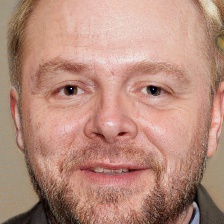}
    \end{tabular}
    }
    \caption{Ablation study. We show inversion results with only ID loss, ID and landmark losses and all three proposed losses.}
    \label{fig:ablation}
\end{figure}

\para{Image reconstruction}
We now pose the image reconstruction problem as optimising the weighted sum of the ID, landmark and histogram losses:
\begin{equation}
    \min_{\mathbf{z}} w_1L_{\text{ID}}(\mathbf{z}) + w_2L_{\text{landmarks}}(\mathbf{z}) + w_3L_{\text{histogram}}(\mathbf{z}),
\end{equation}
where we use $w_1=1$, $w_2=0.0006$ and $w_3=0.01$ in our experiments. This process is visualised in Fig.~\ref{fig:descriptor_inversion_with_additional_losses}.


\subsection{Qualitative results}

We now present results of inversion from ID to image. We begin with an ablation study of ID-only inversion in Fig.~\ref{fig:regression_vs_optimising_id_loss}. The results show that ID loss optimisation significantly improves over direct regression. The identities in the bottom row are clearly a better visual match to those in the top row. We then follow with an ablation study of our full inversion pipeline in Fig.~\ref{fig:ablation}. We show input images in the first column and results with various combinations of losses in columns 2-4. We initialise with our ID to GAN latent code regressor. Then we iteratively optimise only ID loss (column 2 - as in Section \ref{sec:IDonlyinv}), ID loss and landmark loss (column 3) and all of ID, histogram and landmark losses (column 4). The result in column 2 convincingly reconstructs the ID of the original person but the pose and lighting are wrong. Introducing landmark loss largely corrects the pose (though note StyleGAN2 is biased towards frontal poses which means large pose angles are often underestimated). Introducing histogram loss yields similar lighting and skin tone producing an image similar to the original.

\begin{figure}[!t]
\centering
\resizebox{\textwidth}{!}{
\begin{tabular}{@{}m{2.5cm}@{\hspace{0.2cm}}c@{\hspace{0.1cm}}c@{\hspace{0.2cm}}c@{\hspace{0.1cm}}c@{\hspace{0.2cm}}c@{\hspace{0.1cm}}c@{}}
 & \multicolumn{2}{c}{{\large Different lighting}} & \multicolumn{2}{c}{{\large Different expression}} & \multicolumn{2}{c}{{\large Different pose}} \\
\vspace{0.1cm} {\large Real image} & 
\raisebox{-.5\height}{\includegraphics[height=0.25\columnwidth]{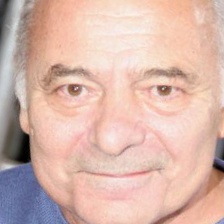}} & \raisebox{-.5\height}{\includegraphics[height=0.25\columnwidth]{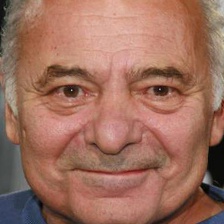}} & \raisebox{-.5\height}{\includegraphics[height=0.25\columnwidth]{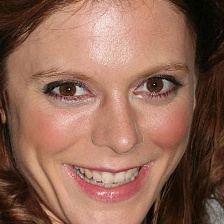}} & \raisebox{-.5\height}{\includegraphics[height=0.25\columnwidth]{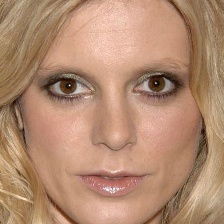}} &
\raisebox{-.5\height}{\includegraphics[height=0.25\columnwidth]{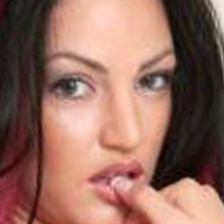}} &
\raisebox{-.5\height}{\includegraphics[height=0.25\columnwidth]{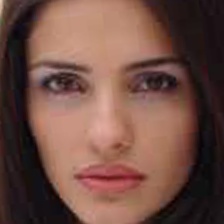}}  \\
{\large Reconstructed from ID} &
\raisebox{-.5\height}{\includegraphics[height=0.25\columnwidth]{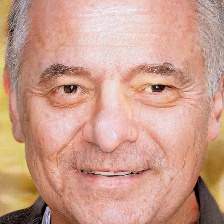}} &
\raisebox{-.5\height}{\includegraphics[height=0.25\columnwidth]{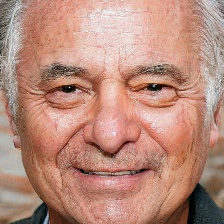}} &
\raisebox{-.5\height}{\includegraphics[height=0.25\columnwidth]{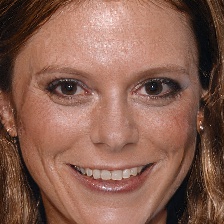}} &
\raisebox{-.5\height}{\includegraphics[height=0.25\columnwidth]{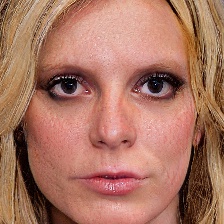}} &
\raisebox{-.5\height}{\includegraphics[height=0.25\columnwidth]{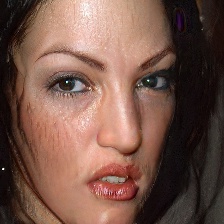}} &
\raisebox{-.5\height}{\includegraphics[height=0.25\columnwidth]{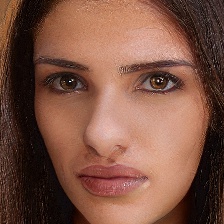}} \\
\end{tabular}
}
\caption{Reconstruction of the same person twice under very different lighting/pose/expression. For each case, we show the original images for the same person (i.e.~same ID) in the first row and the corresponding reconstructions in the second row.}
\label{fig:qualitative_reconstructions_same_id_different_setting}
\end{figure}

Next, we illustrate that our approach is capable of reconstructing different images of the same person under different conditions. In Fig.~\ref{fig:qualitative_reconstructions_same_id_different_setting} we show pairs of real images of the same person in row one. From left to right, these exhibit different lighting, expression and pose. We show our full inversion result in row two. Even though both original images should yield the same ID descriptor, there is enough leaked information that we are able to convincingly reconstruct lighting, expression and pose.

\begin{figure}[!t]
    \centering
    \resizebox{\textwidth}{!}{
    \begin{tabular}{@{}c@{\hspace{0.05cm}}c@{\hspace{0.05cm}}c@{\hspace{0.05cm}}c@{\hspace{0.05cm}}c@{\hspace{0.05cm}}c@{\hspace{0.05cm}}c@{}}
        \rotatebox[origin=l]{90}{\hspace{0.9cm} Target} & 
        \includegraphics[height=0.25\columnwidth]{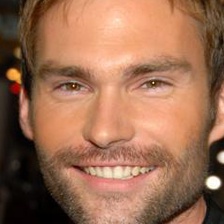} &
        \includegraphics[height=0.25\columnwidth]{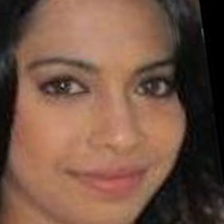} &
        \includegraphics[height=0.25\columnwidth]{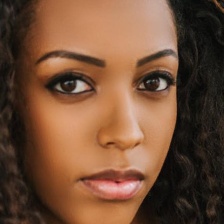} &
        \includegraphics[height=0.25\columnwidth]{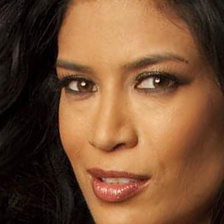} &
        \includegraphics[height=0.25\columnwidth]{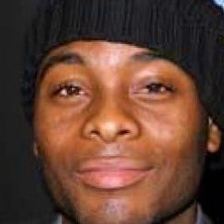} &
        \includegraphics[height=0.25\columnwidth]{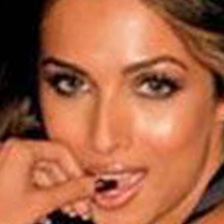}\\
        \rotatebox[origin=l]{90}{\hspace{0.9cm} ID only} & 
        \includegraphics[height=0.25\columnwidth]{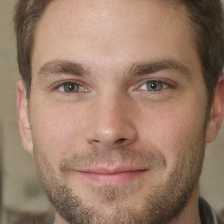} &
        \includegraphics[height=0.25\columnwidth]{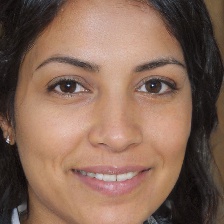} &
        \includegraphics[height=0.25\columnwidth]{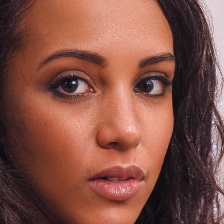} &
        \includegraphics[height=0.25\columnwidth]{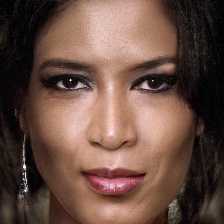} &
        \includegraphics[height=0.25\columnwidth]{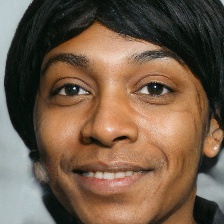} &
        \includegraphics[height=0.25\columnwidth]{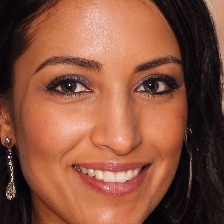} 
        
        \\
        \rotatebox[origin=l]{90}{\hspace{0.4cm} Final Result} & 
        \includegraphics[height=0.25\columnwidth]{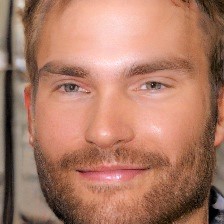} &
        \includegraphics[height=0.25\columnwidth]{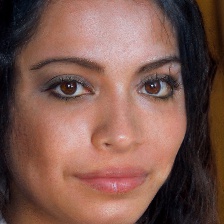} &
        \includegraphics[height=0.25\columnwidth]{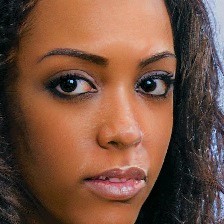} &
        \includegraphics[height=0.25\columnwidth]{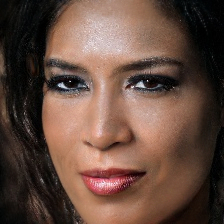} &
        \includegraphics[height=0.25\columnwidth]{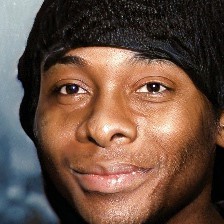} &
        \includegraphics[height=0.25\columnwidth]{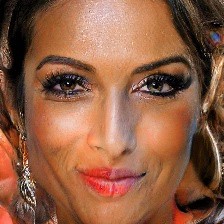} \\
    \end{tabular}
    }
    \caption{Additional inversion results. We show original target image (top row), reconstructions using only ID loss (middle) and full reconstruction result (bottom). The last column shows a failure case.}
    \label{fig:invresults}
\end{figure}

Finally, we show additional inversion results in Fig.~\ref{fig:invresults}. The last column shows a failure case in which the pose is incorrectly reconstructed. This occurs when estimated landmark accuracy is low and is further compounded by the StyleGAN2 bias towards frontal faces.

\subsection{Quantitative results}

We quantitatively evaluate the reconstructed images, comparing them to the original images. To facilitate that, we calculate the mean squared error (MSE), peak signal-to-noise ratio (PSNR) and structural similarity index (SSIM)~\cite{wang2004SSIM} between each reconstructed image and the original image, on the MoFA-Test dataset~\cite{DBLP:conf/iccv/TewariZK0BPT17}, containing 84 images and 78 identities. Table~\ref{tab:MOFA-test} shows an ablation study of reconstructing only with ID loss, with ID and landmark loss, and with all losses. It can be seen that overall, the extra losses help to recreate the actual image and not just the identity of a person.

\begin{figure}[!t]
    \centering
    \includegraphics[width=\columnwidth,trim={30 0 0 0},clip]{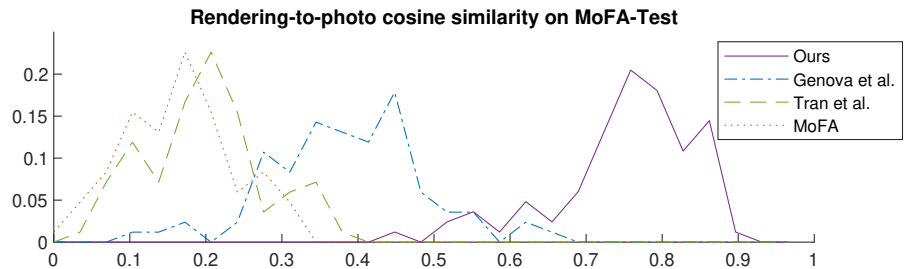} 
    \caption{Distribution of VGGFace cosine similarity for MoFA-Test. We show the distribution of similarity scores of our method, Genova et al.~\cite{genova2018unsupervised}, Tran et al.~\cite{DBLP:conf/cvpr/TranHMM17}, and MoFA~\cite{DBLP:conf/iccv/TewariZK0BPT17} for the original images and their corresponding reconstruction.}
    \label{fig:mofa_test_evaluation}
\end{figure}

Second, we test how well the reconstructed images (using all losses) preserve identity on the MoFA-Test dataset. We use cosine similarity on VGGFace \cite{parkhi2015deep} (as opposed to VGGFace2, which is used for our inversion) embeddings as a measure of how well a method was able to reconstruct the identity. 
Fig.~\ref{fig:mofa_test_evaluation} shows the distribution of similarity scores of our method, compared with Genova et al.~\cite{genova2018unsupervised}, Tran et al.~\cite{DBLP:conf/cvpr/TranHMM17}, and MoFA~\cite{DBLP:conf/iccv/TewariZK0BPT17}. Note that these three methods solve a different problem: reconstruction with a 3D morphable model \emph{given the original image}. However, Genova et al.~\cite{genova2018unsupervised} do this via an ID bottleneck meaning the comparison is meaningful. With an average similarity score of $0.77$, we significantly outperform all other methods ($0.40$ for Genova et al., $0.22$ for Tran et al., and $0.18$ for MoFA). This is particularly notable given that we reconstruct the image \emph{only from an ID vector}. The difference is likely partly down to using a generative model (StyleGAN2) that is much more powerful than a 3DMM.

\begin{table}[!t]
\centering
\begin{tabular}{@{}l@{\hspace{0.4cm}}c@{\hspace{0.4cm}}c@{\hspace{0.4cm}}c@{}}
\toprule
     & ID only    & ID+LMs    & ID+LMs+Hist \\
     \midrule
MSE (lower better)  & 2.05  & 2.14  & 2.03    \\ 
PSNR (higher better) & 12.98  & 13.21   & 13.35      \\
SSIM (higher better) & 0.14 & 0.14 & 0.15   \\
\bottomrule
\end{tabular}
\vspace{0.4em} 
\caption{Quantitative evaluation on MOFA-test, comparing reconstruction with only ID loss, ID and landmark loss, and ID, landmark and histogram loss.}
\label{tab:MOFA-test}
\end{table}

\section{Conclusion}
Our results show that, indeed, non-identity information finds its way into state-of-the-art face descriptor embedding networks like VGGFace2 and ArcFace. We have shown that, given the possibility of being able to query the network, it is possible to not only reconstruct an image of a person's face encoded in a descriptor but also non-ID attributes as well as landmark positions and the image histogram. Being able to reconstruct not just an image with the right ID but the actual original input image has privacy and security implications.

There are many important avenues for future work. First, it is important to replicate these results on other face embedding networks (our initial experiments suggest that our findings indeed transfer between networks). Second, the inversion performance could likely be improved by including other explicit non-ID features. Thirdly, our current work only tried to inversion the background colour by predicting the histogram of RGB intensity from ID, and the subsequent work could be extended by restoring the solid objects on the background to provide a more intuitive view of the background information leaks.
Finally, and most interestingly, we believe that our work provides a route to improving face recognition performance while also alleviating privacy concerns. Non-ID information is a nuisance factor for face recognition. It means that some of the capacity of the embedding space is wasted on useless information and that distance measures incorrectly observe identity dissimilarity when in fact the difference is due to non-ID factors. This is in addition to privacy and security concerns related to the leakage of image information into ID vectors. In future we will introduce an adversarial loss into face recognition training that penalises the inclusion of non-ID information in the embedding. We will also attempt to reconstruct non-id information from black-box features, without access to the model’s architecture. 



%
%
%
\bibliographystyle{splncs04}
\bibliography{refs}



\end{document}